\documentclass[reprint,amsmath,amssymb,aps,prl,superscriptaddress]{revtex4-2}
\usepackage[utf8]{inputenc}
\usepackage[colorlinks]{hyperref}
\usepackage{graphicx}
\usepackage{dcolumn}
\usepackage{bm}
\usepackage[space]{grffile}
\usepackage[export]{adjustbox} 
\usepackage{booktabs}
\usepackage[caption=false]{subfig}

\newcommand{\henon}{H{\'e}non}

\begin{document}

\title{
Machine-learning optimized measurements of chaotic dynamical systems via the information bottleneck
} 

\author{Kieran A. Murphy}
\affiliation{Dept. of Bioengineering, School of Engineering \& Applied Science,}
\author{Dani S. Bassett} 
\affiliation{Dept. of Bioengineering, School of Engineering \& Applied Science,}
\affiliation{Dept. of Electrical \& Systems Engineering, School of Engineering \& Applied Science,}
\affiliation{Dept. of Neurology, Perelman School of Medicine,}
\affiliation{Dept. of Psychiatry, Perelman School of Medicine,}
\affiliation{Dept. of Physics \& Astronomy, College of Arts \& Sciences, University of Pennsylvania, Philadelphia, PA 19104, USA}

\affiliation{The Santa Fe Institute, Santa Fe, NM 87501, USA}

\begin{abstract}
Deterministic chaos permits a precise notion of a “perfect measurement” as one that, when obtained repeatedly, captures all of the information created by the system’s evolution with minimal redundancy.
Finding an optimal measurement is challenging, and has generally required intimate knowledge of the dynamics in the few cases where it has been done.  
We establish an equivalence between a perfect measurement and 
a variant of the information bottleneck.
As a consequence, we can employ machine learning to optimize measurement processes that efficiently extract information from trajectory data.
We obtain approximately optimal measurements for multiple chaotic maps and lay the necessary groundwork for efficient information extraction from general time series.
\end{abstract}

\maketitle

Encapsulated in deterministic chaos is the fundamental obstruction to predictability that can result from nonlinearity in a system's evolution, even in the absence of randomness~\cite{shaw1981,eckmannruelle1985}.
Signatures of chaos are found broadly, from weather~\cite{tsonis1989chaosweather,slingo2011weatheruncertainty} to the brain~\cite{kargarnovin2023evidence,skinner1992application}, and tools developed in the study of chaos have been applied more broadly still~\cite{nicolis2012foundations,bradley2015nonlinear}.
Advancing capabilities to forecast chaotic dynamics thus has marked potential for impact.
The challenge may be glimpsed through the relation between precision and predictability: for any predictive model utilizing less than infinite precision, the error of prediction grows exponentially~\cite{wales1991calculating}. 
Given the inherent difficulties and the potential for impact, the field has recently turned to machine learning~\cite{pathak2018model,amil2019machine,tang2020introduction,gilpin2023forecast}.

Machine learning approaches to forecasting chaotic dynamics generically utilize full-precision states as input to the predictive model. 
Yet, the elusive determinism of chaos gives rise to a curious fact: beyond a certain precision per state, the ability to forecast given a partial trajectory saturates~\cite{shaw1981}.
There is thus a measurement capacity beyond which resources---whether to acquire the measurement, or to record the trajectory---are wasted.
Instead, it is sufficient to discretize the continuous-valued states, yielding an analogous system with \textit{symbolic dynamics} that simplifies the statistical analysis of the system ~\cite{williams2004introduction,nicolis2012foundations,beal2013symbolic,lind2021symbolicdyn} and can be used for various applications~\cite{hirata2023review} including anomaly detection~\cite{ray2004symbolic} and communication~\cite{hayes1994experimental}.
Here we employ machine learning to optimize the measurement of, and equivalently the extraction of information from, a chaotic system.

The requisite precision per measurement, in the form of a number of bits per state, is a fundamental quantity of the system: the metric entropy~\cite{eckmannruelle1985}.
Also known as the Kolmogorov-Sinai (KS) entropy, it corresponds to the rate of information creation, generated from infinitesimal scales by the expansion of nearby points under the dynamics and commonly referred to as sensitive dependence on initial conditions~\cite{shaw1981,james2014chaosforgets}.
For many systems of interest, the metric entropy is equal to the sum of the system's positive Lyapunov exponents~\cite{eckmannruelle1985}.

\begin{figure}
    \includegraphics[width=\columnwidth]{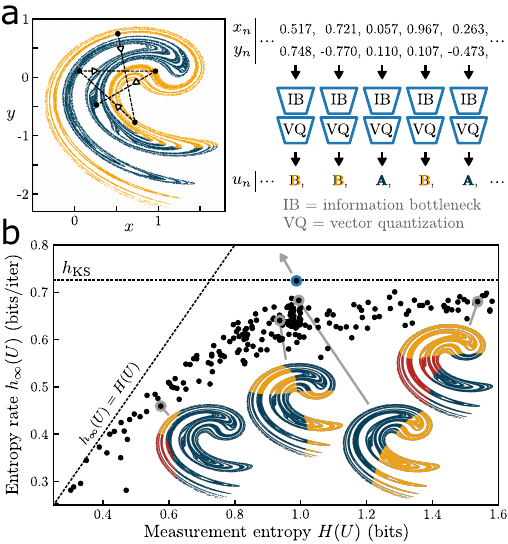}
    \caption{\textbf{(a)} The attractor of the Ikeda map and a partial trajectory.
    States colored blue (orange) are measured as outcome \textit{A} (\textit{B}).
    The proposed method optimizes the measurement of a chaotic system using artificial neural networks that apply an information bottleneck (IB) and vector quantization (VQ) to each continuous-valued state.
    \textbf{(b)} The space of possible discrete measurements visualized in terms of the entropy of a single measurement, $H(U)$, and the rate of entropy production in the infinite duration limit, $h_\infty(U)$.
    The entropy rate of any partition is upper bounded by $H(U)$ and the metric entropy $h_\textnormal{KS}$ (dashed lines).
    The blue point corresponds to the partition in panel \textbf{a} that was optimized with the proposed method.
    The black points correspond to measurements parameterized by neural networks with random weights. 
    Error bars on the entropy rates are within the markers.
    }
    \label{fig:partitions}
\end{figure}

A finite metric entropy---which can be used to define chaos~\cite{gaspard1993noise,beck1995thermodynamics,schurmanngrassberger1996} and quantify the extent of chaos in a system~\cite{cohenprocaccia1985,wales1991calculating}---implies that a system's continuous-valued trajectory through state space has the same information content, in the asymptotic limit, as a corresponding sequence of discrete-valued measurements, although only if the measurement process is optimal.
Discrete-valued measurements \textit{color} state space according to a partition, clustering states according to the partition element to which they belong.
Optimizing over the enormous space of possible partitions has traditionally been avoided by finding special partitions called generators (or generating partitions) that are known to extract all information created by the dynamics~\cite{badii1997complexity}.
Generators may be remarkably coarse~\cite{kennelbuhl2003gp}; one approximate generator for the Ikeda map~\cite{ikeda1979} requires only two colors to partition state space (Fig.~\ref{fig:partitions}a).
Despite the intricate structure of the attractor, a measurement of the state with the capacity of one bit captures all information created by the dynamics.

Finding a generator is challenging and has been accomplished for only a limited number of low-dimensional systems.
Traditionally, methods have leveraged intimate knowledge of the dynamics~\cite{grassberger1985henon,jaegerkantz1997,davidchack2000UPOs,plumecoq2000templateanalysisII,mitchell2012partitioning,chai2021symbolic,zhang2022koopman}, reflecting the fundamental connection between optimal information extraction and structure in the system's state space. 
As an alternative, purely data-driven approaches can reveal structure in the dynamical system with increased robustness to noise and the potential to scale to higher dimensions~\cite{kennelbuhl2003gp,hirata2004shadowing,patil2018empirical,ghalyan2018locally,rubido2018entropy}.
Whereas previous data-driven approaches have utilized classical methods of clustering with heuristics to simplify the search problem,
here we bring the expressivity of deep learning to the task of optimizing information extraction from a chaotic system.
Our focus is on partitions that perfectly extract information, of which generators are a subset~\cite{eckmannruelle1985}, though in practice we find that the optimized partitions are closely related to previously established generators.
Central to our approach is an equivalence between efficient information extraction and an objective from rate-distortion theory known as the distributed information bottleneck~\cite{aguerri2018DIB}.

The distributed information bottleneck is a rate-distortion scenario to optimize the lossy compression of multiple sources of information individually so as to maximize collective information about an auxiliary quantity~\cite{aguerri2018DIB,dib_ml}.
The lossy compression extracts some information and discards the rest, and is accomplished by distributing an information bottleneck (IB)~\cite{tishbyIB2000} to each source.
By extracting important bits of information across multiple sources, the distributed IB has been used to decompose information in complex systems~\cite{dib2} and to provide interpretability to black-box machine learning models by identifying the information utilized for prediction~\cite{dib_ml}.
Here we use the distributed IB to lossily compress each state in a finite trajectory such that the sequence of measurements contains maximal information about a ``reference'' state taken from the trajectory.

Let a state $x \in \mathcal{M}$ exist in $\mathbb{R}^d$, where $d$ is the dimension of the state space.
A map $\mathcal{F}(x): \mathcal{M} \rightarrow \mathcal{M}$ propagates a state forward in time by one iteration, i.e. $\mathcal{F}(x_n) = x_{n+1}$.
We consider discrete time maps in this work; for the purposes of extracting information, a continuous-time flow can be converted to a discrete map by sampling with a fixed interval~\cite{eckmannruelle1985}.
The dynamics are fully described by the map $\mathcal{F}$, but a probabilistic view is often more natural~\cite{nicolis2012foundations} and allows us to utilize information theory.
For ergodic dynamical systems, which will be our concern in this work, the natural probability distribution over states $p(x)$ is an invariant measure, $\mathcal{F}(p(x)) = p(x)$, and can be obtained by iterating forward a long trajectory~\cite{eckmannruelle1985}.

Given a probability distribution over states, we can define a random variable $X_n$ for the state at timestep $n$, and a random process $\boldsymbol{X}_{n:n+L}$ for a sequence of random variables $X_n, X_{n+1}, ..., X_{n+L-1}$.
For a stationary process, the start of the trajectory is unimportant, and instead we consider the statistics of subsequences of length $L$, which we denote $\boldsymbol{X}_L$.

A continuous-valued state can be \textit{measured}, and specifically, discretized, through the use of a partition that divides the support of $p(x)$ into disjoint subsets~\cite{kennelbuhl2003gp}.
The outcome of a measurement, a random variable $U=f(X)$, converts the continuous-valued state to the index of the subset to which it belongs.
Thus a sequence $\boldsymbol{X}_L$ is replaced by a sequence of discrete measurements $\boldsymbol{U}_L$.

How much information does a measurement $U$ convey about the original state $X$?
Intuitively, information gained reduces uncertainty.
The amount of uncertainty about the outcome of a random variable $Z$ may be quantified via the Shannon entropy, $H(Z)=\mathbb{E}_{z\sim p(z)}[-\textnormal{log} \ p(z)]$~\cite{shannon1948mathematical}.
The mutual information contained in two random variables is given by the reduction of entropy in one variable after finding the value of the other, $I(Z_1;Z_2)=H(Z_1)-H(Z_1|Z_2)$~\cite{cover1999elements}.
Measuring a state---by recording in which subset of a partition it resides---conveys $I(U;X)=H(U)$ bits of information because the mapping from $X$ to $U$ is deterministic (i.e., $H(U|X)=0$).

As a random process plays out, entropy is generated from the uncertainty about each subsequent outcome. 
The entropy rate is defined as the average entropy generated per step in the limit where the sequence length becomes infinite, $h_\infty(U)=\lim_{L\rightarrow \infty} H(\boldsymbol{U}_L)/L$~\cite{cover1999elements}.
The largest achievable entropy rate of any partition $U$ is the KS entropy~\cite{eckmannruelle1985},
\begin{equation}
    h_\textnormal{KS} = \sup_U h_\infty(U).
\label{eqn:h_ks}
\end{equation}
A ``trivial'' partition that achieves the maximal entropy rate can be approximated by a fine discretization of state space~\cite{cohenprocaccia1985}, though it is possible for partitions to capture all information produced by the dynamics and also be coarse, i.e. with minimal entropy~\cite{kennelbuhl2003gp} (Fig.~\ref{fig:partitions}a).

The objective for a partition with minimal entropy and maximal entropy rate can be cast as a rate-distortion problem~\cite{cover1999elements}.
Viewing the measurement as a communication channel, we simultaneously minimize the information transmitted about a single state $I(U;X)$ while maximizing the information transmitted about the trajectory, $I(\boldsymbol{U}_L;\boldsymbol{X}_L)$.
For a chaotic system, a single continuous-valued state contains the same information as any portion of its trajectory, $I(\boldsymbol{U}_L;X_\text{ref})=I(\boldsymbol{U}_L;\boldsymbol{X}_L)$.
By using a large but finite $L$, we recover a distributed IB formulation where the states in a sequence serve as the distributed sources of information and the reference state is the auxiliary variable. 
We minimize the distributed IB Lagrangian~\cite{aguerri2018DIB,dib_ml} over possible measurements $U$,
\begin{equation}
    \mathcal{L} = - I(\boldsymbol{U}_L;X_\text{ref}) + \beta \sum_{i=0}^{L-1} I(U_i;X_i),
\label{eqn:dib}
\end{equation}
where $\beta$ is a parameter that determines the cost of information transmission relative to information about $X_\text{ref}$.

\begin{figure}
    \centering
    \includegraphics[width=\columnwidth]{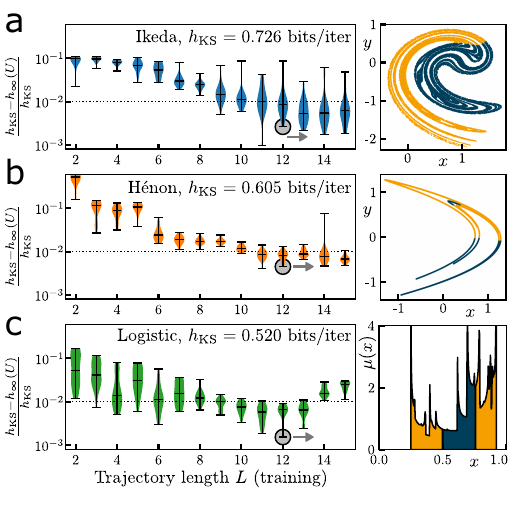}
    \caption{
    The fractional deviation of the entropy rate $h_\infty(U)$ from the metric entropy $h_\text{KS}$ as a function of the length of the trajectory $L$ used for training for \textbf{(a)} Ikeda and 
    \textbf{(b)} \henon\ maps with standard parameters, and \textbf{(c)} the logistic map with $r=3.7115$. 
    The distribution of values of the entropy rate over 20 trials for each $L$ are shown as violin plots, with the extrema and median indicated by the black horizontal marks; the dotted line indicates $h_\infty(U)=0.99h_\text{KS}$.
    For each system, the partition found with the largest entropy rate for $L=12$ is shown on the right.
    For the logistic map, the probability density of states $p(x)$ is colored according to the optimized partition.
    }
    \label{fig:sequence_length_dependence}
\end{figure}

\begin{figure*}
    \centering
    \includegraphics[width=\linewidth]{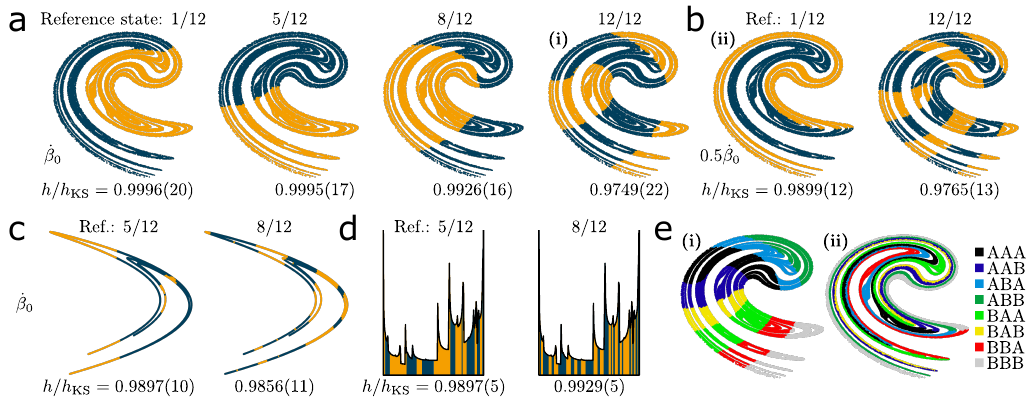}
    \caption{ 
    \textbf{(a)} For optimized partitions of the Ikeda map, trained with sequences of length $L=12$ and with a base information annealing rate $\dot{\beta}=\dot{\beta}_0$, the reference state (i.e., the one predicted from the sequence of measurements) strongly influences the found partition.  
    Each displayed partition is the one with the largest entropy rate (listed under the partition) after ten trials.
    Coordinate axes have been suppressed.
    \textbf{(b)} The same as in panel \textbf{a}, but with half the rate of information annealing $\dot{\beta}=0.5\dot{\beta}_0$ during training.
    \textbf{(c)} Two of the partitions of the \henon\ map, identified with the same annealing rate as in panel \textbf{a}.
    \textbf{(d)} Same as in panel \textbf{c} for the logistic map with $r=3.7115$.  
    \textbf{(e)} Colorings of the Ikeda attractor where, given a sequence of three measurements $\boldsymbol{u}_{:3}$, the reference state is the final state $x_3$, utilizing for the measurement $U$ the partition labelled \textbf{(i)} (left) and \textbf{(ii)} (right) in \textbf{a} and \textbf{b}, respectively.
    }
    \label{fig:zoo}
\end{figure*}

The space of possible partitions is vast, filled with colorings of the attractor that encapsulate suboptimal information~\cite{bollt2001misplaced,cafaro2015causation}.
Fig.~\ref{fig:partitions}b displays example partitions of the Ikeda map by their measurement entropy $H(U)$ and entropy rate $h_\infty(U)$.
By replacing the mutual information terms in Eqn.~(\ref{eqn:dib}) with appropriate bounds---namely InfoNCE~\cite{oord2018InfoNCE} as a lower bound on $I(\boldsymbol{U}_L;X_\text{ref})$ and the variational upper bound central to variational autoencoders on $I(U_i;X_i)$~\cite{betavae,alemiVIB2016,aguerriDVIB2021}---we can train artificial neural networks that parameterize the space of partitions.

The lossy compression that maps each state $x$ to its corresponding measurement $u$ presents a difficulty for deep learning because quantization operations are non-differentiable~\cite{jang2016categorical}.
By contrast to \textit{hard} measurements, where $H(U|X)=0$, \textit{soft} measurements contain stochasticity that can be used to facilitate optimization~\cite{agustsson2017softtohard}.
We trained with soft measurements that were hardened for inference, and used a two-step compression process to gradually remove information about the state. 
Each step was performed by a distinct multilayer perceptron (MLP) (Fig.~\ref{fig:partitions}a).
The first MLP performed a soft measurement by mapping $x$ to a distribution, $p(\tilde{u}|x)$, in an intermediate representation space where the Kullback-Leibler divergence penalty from variational autoencoders bottlenecked the transmitted information~\cite{betavae,alemiVIB2016}.
A sample from the distribution $p(\tilde{u}|x)$ was then input to a second MLP---the vector quantizer---whose output could be smoothly transitioned from a soft to a hard measurement.
The output $u$ was a probability vector over symbols (e.g., \textit{A} and \textit{B} in Fig.~\ref{fig:partitions}a) parameterized by a temperature-like parameter that discretizes the assignment in one limit by making all probability distributions a one-hot vector~\cite{agustsson2017softtohard}.
The magnitude of the information bottleneck penalty $\beta$ was annealed during training to gradually increase information transmission.
For learning binary partitions of the systems studied in this work, training was stopped after the information transmitted by the information bottleneck (the first MLP) exceeded one bit. 

All states $\boldsymbol{x}_{L}$ were `measured' and then the corresponding measurements $\boldsymbol{u}_{L}$ were input to a third MLP that predicted the reference state $x_\text{ref}$.
To directly maximize information extraction $I(\boldsymbol{U}_L;X_\text{ref})$, the InfoNCE loss~\cite{oord2018InfoNCE} evaluates the prediction in a shared representation space to which both $\boldsymbol{u}_L$ and $x_\text{ref}$ are mapped (requiring a fourth MLP that maps $x_\text{ref}$ to the space). 
The loss quantifies how similar the representation for the sequence $\boldsymbol{u}_L$ is to that of its corresponding $x_\text{ref}$, in comparison to the representations of a batch of states sampled randomly from the entire attractor.
Network architectures and training details may be found in the Supp.

To assess optimized measurements, the difference between the known metric entropy and the measurement's entropy rate gives the information per iteration that the measurement fails to capture.
Lossless data compression commonly forms the basis of approaches to estimate entropy rate because the same regularities in a sequence that reduce its entropy rate are what data compression is designed to leverage to shrink file size~\cite{schurmanngrassberger1996,kennel2002ctw,gao2008estimating,martiniani2019quantifying}.
We compared several methods on symbolized sequences from chaotic systems with known generators (Supp.) and obtained the most precise estimates with a form of data compression known as context tree weighting~\cite{willems1995ctw}, whereby we extrapolated from finite size scaling with an ansatz proposed in \citet{schurmanngrassberger1996}. 

While strict equivalence between the distributed IB and a coarse partition with entropy rate equal to $h_\text{KS}$ is valid only in the limit where the sequence length $L$ is infinite, we found that $L\approx10$ was sufficient to consistently find partitions with an entropy rate at least $0.99 h_\text{KS}$ for the chaotic maps studied (Fig.~\ref{fig:sequence_length_dependence}).
Optimization was robust, with the performance of twenty trials tightly clustered for larger $L$ (displayed as the violin plots in Fig.~\ref{fig:sequence_length_dependence}).  
Partitions with the largest entropy rate for $L=12$ are displayed in the right column of Fig.~\ref{fig:sequence_length_dependence}.
For the \henon\ and logistic maps, the optimized partitions are slight variations of reported generators~\cite{grassberger1985henon,schurmanngrassberger1996}, while the partition for the Ikeda map differs from generators reported in prior work~\cite{davidchack2000UPOs,kennelbuhl2003gp,hirata2004shadowing,ghalyan2018locally}.

The generator for a chaotic map is not unique~\cite{jaegerkantz1997}.
All images and preimages of a generating partition (i.e., iterating the points in each partition element forward or backward in time while maintaining the assignments) are also generators.
Additionally there can be nontrivial variants that cross through special points called homoclinic tangencies, where the stable and unstable manifolds run parallel to one another~\cite{jaegerkantz1997}.
We found that certain details of the training process steered optimization to qualitatively different partitions (Fig.~\ref{fig:zoo}).
Varying the reference state drove the optimization to different iterates of the same base partition (Fig.~\ref{fig:zoo}a-d; iterates in Supp.), and a slower annealing of the information $\dot{\beta}$ reached deeper into the forward and backward iterates of the base partition (Fig.~\ref{fig:zoo}b).
Different iterates require different functions to integrate multiple measurements (Fig.~\ref{fig:zoo}e); the iterates found through optimization allow for simpler integration functions.
The base partition found for the Ikeda map (Fig.~\ref{fig:zoo}a, reference state 5 of $L=12$) closely resembles what has been found in prior work~\cite{davidchack2000UPOs}.
 
In this Letter, we established an equivalence between the distributed information bottleneck and the minimal partition able to capture all information generated by a chaotic system, and leveraged the equivalence to optimize measurement of a chaotic dynamical system with machine learning.
Operating without knowledge of the dynamics and without reliance on properties specific to generating partitions such as the unique description of periodic orbits~\cite{davidchack2000UPOs}, homoclinic tangencies~\cite{grassberger1985henon,jaegerkantz1997,mitchell2012partitioning,chai2021symbolic}, or the Koopman operator~\cite{zhang2022koopman}, the method is not restricted to chaotic systems.
Instead, the deterministic chaos serves as a testbed where the notion of an optimal measurement scheme has been made precise and can be quantitatively evaluated.
Optimizing a measurement to efficiently extract maximal information about an underlying state is a broad goal in understanding inference in biological and computational systems.
Prior research has predominantly focused on a singular compression that extracts maximal information from the present and/or past about the future~\cite{creutzig2009PIB,still2010optimal,palmer2015predictive,marzen2016predictive}.
A notable exception is the recursive information bottleneck~\cite{still2014information}, in which a sequence of measurements are recursively aggregated and each step's measurement scheme can vary based on what has been previously observed.
By contrast, the distributed information bottleneck setup of this work optimizes a fixed measurement scheme that is repeatedly applied for maximal aggregate information, a plausible scenario for sensing organisms or sensory devices.

The current study focused exclusively on relatively simple chaotic maps that have been well characterized so as to establish the capabilities of the proposed method.  
The variegated faces of chaos present exciting opportunities for the optimization of measurement processes and
can serve as a rich testbed for machine learning methods of compression~\cite{jang2016categorical,gilpin2023forecast}.
In the large majority of systems where there is no precedent partition with which to compare, the metric entropy can be readily estimated from data~\cite{cohenprocaccia1985,wales1991calculating} and other means of quantitatively evaluating partitions can be used~\cite{plumecoq2000templateanalysisII} to ground the measurement schemes learned by the proposed method.

To our knowledge, the connection between metric entropy and a rate-distortion objective was previously unconsidered in the literature.
The transformation $\mathcal{F}$ whose repeated application creates the strange attractor evolves points in state space such that a fixed measurement can continually acquire new information about the continuous-valued trajectory \textit{ad infinitum}.
By optimizing the minimally redundant lossy compression of the attractor, the spawned information that serves to limit predictability for any finite model is manifest as a specific coloring of the attractor that divides state space with a remarkably crude, yet highly specific cut. 

\section{Acknowledgements}
We gratefully acknowledge Sam Dillavou, Sarah E. Marzen, Kevin A. Mitchell, and Navendu S. Patil for helpful discussions, and Jason Z. Kim, and Suman Kulkarni for comments on the manuscript.

\clearpage

\onecolumngrid

\setcounter{section}{0}
\setcounter{page}{1}
\setcounter{figure}{0}
\setcounter{table}{0}

\renewcommand{\thepage}{S\arabic{page}}
\renewcommand{\thesection}{S\arabic{section}}
\renewcommand{\thetable}{S\arabic{table}}
\renewcommand{\thefigure}{S\arabic{figure}}
\renewcommand{\figurename}{Supplemental Material, Figure}

\hrule
\vspace{3mm}

{\Large Supplemental Material }

\section{Code availability}
The full code base has been released on Github at the following location: \href{https://distributed-information-bottleneck.github.io}{distributed-information-bottleneck.github.io}.
Every analysis included in this work can be repeated from scratch in Google Colab with the iPython notebook in \href{https://github.com/distributed-information-bottleneck/distributed-information-bottleneck.github.io/tree/main/chaos}{this directory}.

\section{Appendix A: The chaotic maps studied}
We studied three chaotic maps in this work.
The logistic map~\cite{beck1995thermodynamics},
\begin{equation}
    x_{n+1} = r x_n (1-x_n),
\end{equation}
is a one-dimensional map defined over the unit interval.  
We used the value $r=3.7115$ for the main text, and additionally used $r=4$ to benchmark the entropy rate estimation in App. B.
The \henon\ map~\cite{grassberger1985henon},
\begin{equation}
    \begin{split}
        x_{n+1} &= 1 - ax_n^2 + by_n\\
        y_{n+1} &= x_n,
    \end{split}
\end{equation}
is a two-dimensional map, and we used the extensively studied parameters $a=1.4$, $b=0.3$.
Finally, the Ikeda map~\cite{ikeda1979},
\begin{equation}
    \begin{split}
        \phi &= \kappa - \eta / (1 + x_n^2 + y_n^2)\\
        x_{n+1} &= a + b(x_n \cos(\phi) - y_n \sin(\phi))\\
        y_{n+1} &=  b(x_n \sin(\phi) - y_n \cos(\phi)),
    \end{split}
\end{equation}
is another two-dimensional map, and we again followed precedent with the parameters $a=1$, $b=0.9$, $\kappa=0.4$, $\eta=6$~\cite{davidchack2000UPOs}.

\section{Appendix B: Estimation of entropy rate}
Given a sequence of symbols that can take any of a discrete set of values, we wish to estimate the asymptotic rate of entropy produced per symbol,
\begin{equation}
    h_\infty(U) = \lim_{L \rightarrow \infty} \frac{H(\boldsymbol{U}_L)}{L} = \lim_{L \rightarrow \infty} H(U_L|\boldsymbol{U}_{L-1}).
\end{equation}
The second equality holds in the case of stationary processes~\cite{gao2008estimating} and makes clearer the interpretation of the entropy rate as the rate of information creation.

The entropy rate can be at most $H(U)$, when every symbol is independent from the rest, as is the case for a sequence of coin flips.
Regularity in the process reduces the entropy rate, and the lowest entropy rate of zero corresponds to determinism in the symbolic dynamics (e.g., a periodic signal).  
Note the distinction between determinism in the symbolic dynamics, where every symbol is known with certainty given the past, and determinism in the continuous-valued state space.  
The latter can yield uncertainty given only the results of discrete measurements, which have discarded information.

\begin{figure*}
    \centering
    \includegraphics[width=\linewidth]{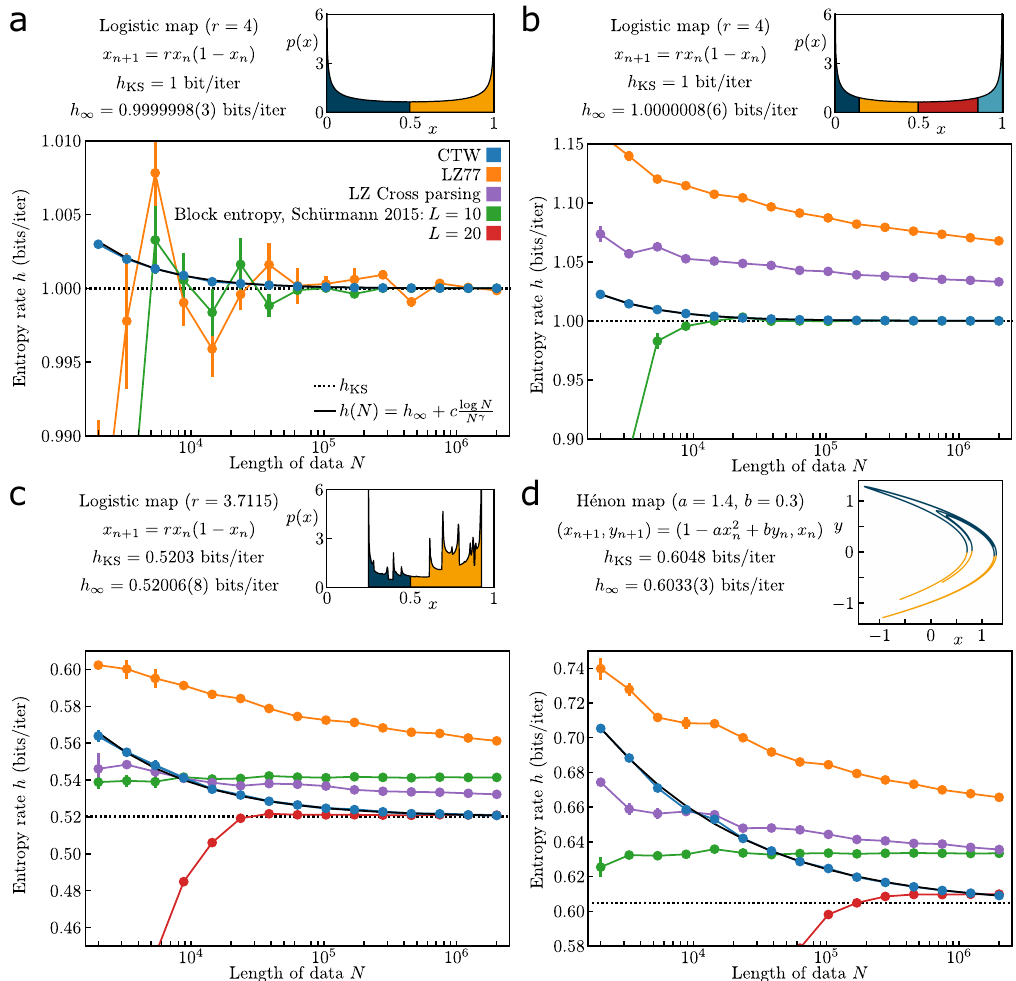}
    \caption{\textbf{Estimating entropy rate.}
    \textbf{(a)} The logistic map ($r=4$) generating partition with two symbols; $h_\textnormal{KS} = 1$ bit per iteration of the map.
    \textbf{(b)} The logistic map ($r=4$) generating partition with four symbols; $h_\textnormal{KS} = 1$ bit per iteration of the map.
    \textbf{(c)} The logistic map ($r=3.7115$) generating partition with two symbols; $h_\textnormal{KS} = 0.5203$ bit per iteration of the map.
    \textbf{(d)} The \henon\ map ($a=1.4, b=0.3$); $h_\textnormal{KS} = 0.6048$ bits per iteration of the map. 
    Error bars indicate the standard error from five repeats.
    }
    \label{fig:entropy_convergence}
\end{figure*}

Methods of file compression leverage regularities in the bit stream of a file, and commonly form the basis for estimates of entropy rate.
Lempel-Ziv (LZ) compression, of which there are several variants, constructs a codebook of previously unobserved sequences during a single pass through the data.
Context tree weighting (CTW) is another method of file compression that constructs a suffix tree storing every subsequence contained in the data.
While LZ compression is fast to compute, the meticulous record keeping of CTW converges more quickly with file size to the entropy rate~\cite{kennel2002ctw}.
We implemented the infinite depth CTW following \citet{kennel2005bayesian}; the C++ code may be found in the linked repository.

We compare in Fig.~\ref{fig:entropy_convergence} various estimates of the entropy rate for generating partitions of the logistic and \henon\ maps where $h_\textnormal{KS}$ and generating partitions are known.
We included a cross parsing variant of LZ whereby a codebook is made with one sequence of length $L$ and evaluated on another~\cite{ro2022xparse}.
Additionally, we compared to a bias-corrected form of block entropy~\cite{schurmann2015blockent}, based on the entropy of subsequences of length $B$.
Even with the bias correction, estimating the entropy rate using block entropy is problematic because there is often no indication about what length $B$ yields the best estimate~\cite{kennel2005bayesian}.
Here and for all other entropy rate estimates in the manuscript, sequences of fifteen lengths $L$ logarithmically spaced between two thousand and two million were randomly selected from a dataset of 20 million points.
Five repeats were evaluated for each $L$.

We found that the convergence of CTW, while faster than LZ, was highly variable for different partitions, and that computing the entropy rate from CTW for a reasonably large file size would not suffice in general.
Instead, we followed \citet{schurmanngrassberger1996}, who fit an ansatz of sequence length dependence of the entropy rate measured via construction of a suffix tree (though not CTW).  
We found the same ansatz to work well for the finite size scaling of the CTW estimates,
\begin{equation}
    h_N = h_\infty + c \frac{\log N} {N^\gamma},
\end{equation}
and used nonlinear least squares (\href{https://docs.scipy.org/doc/scipy/reference/generated/scipy.optimize.curve_fit.html}{\texttt{scipy.optimize.curve\_fit}}) for the estimate of $h_\infty$ and its standard error.

\begin{figure*}
    \centering
    \includegraphics[width=\linewidth]{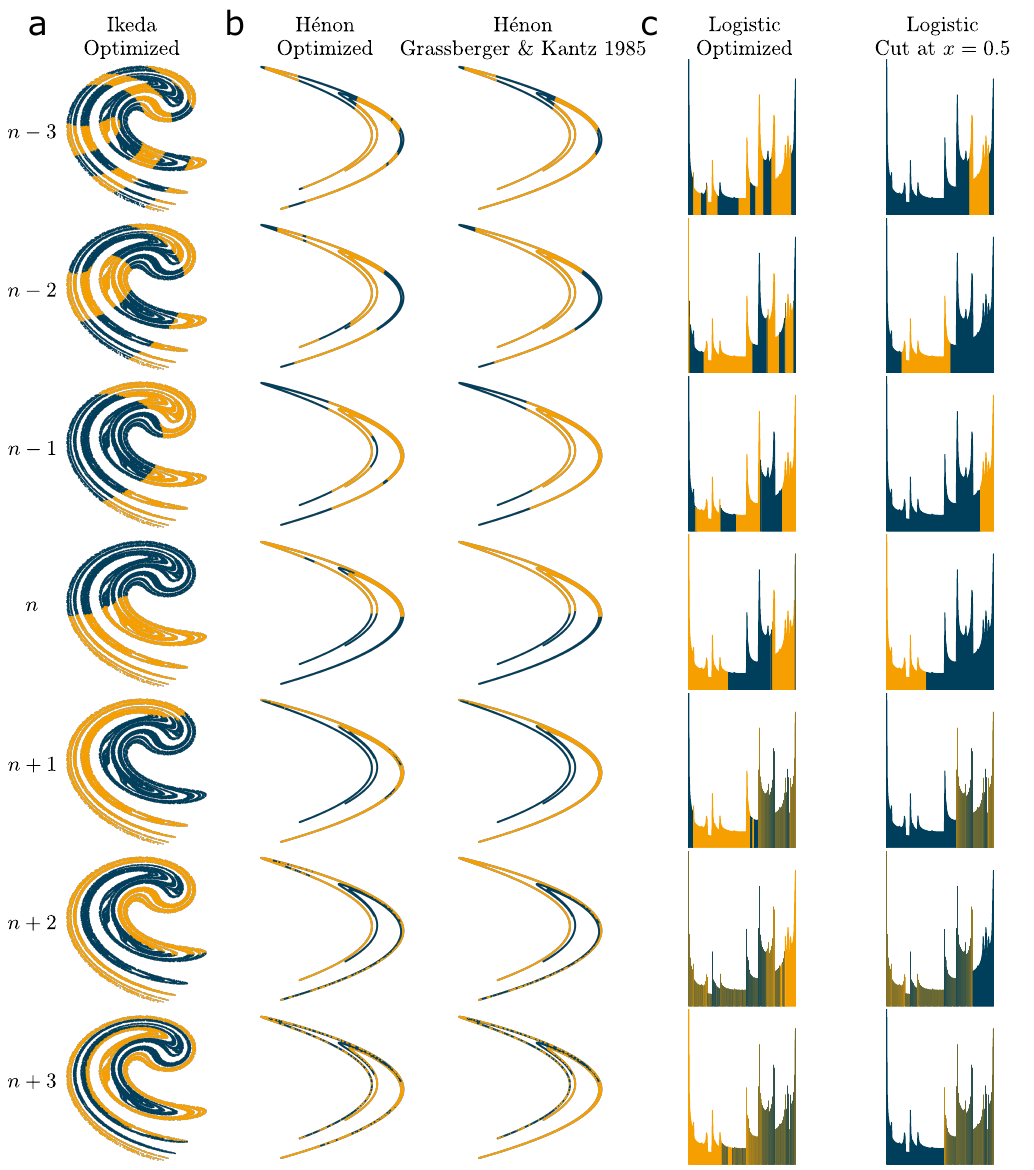}
    \caption{\textbf{Iterates of optimized partitions and established generating partitions.}
    The partitions in the gray box, labelled timestep $n$, were iterated forward and backward three steps while retaining the coloring from timestep $n$.
    Coordinate axes have been suppressed.
    \textbf{(a)} The Ikeda map, with the partition at timestep $n$ the result of a run with $L=12$ and reference timestep $6/12$.  
    \textbf{(b)} The \henon\ map with standard parameters, optimized with the current method on the left and found by \citet{grassberger1985henon} by threading certain homoclinic tangencies.
    \textbf{(c)} The logistic map with $r=3.7115$, optimized with the current method on the left and the known GP, with a boundary at the critical point at $x=0.5$ (true for any value of $r$).
    The invariant measure $p(x)$ is displayed vertically, and the bins are shaded according to the proportion of partition assignments at timestep $n$.
    }
    \label{fig:iterates}
\end{figure*}

\section{Appendix C: Machine learning implementation details}
All experiments were implemented in TensorFlow and run on a single computer with a 12 GB GeForce RTX 3060 GPU.  Each optimization and its entropy rate estimation took several minutes.

For all maps, subsequences of length $L$ iterations were randomly sampled from a sequence of length $10^6$ for training, and every $x_i$ was individually compressed by the same two-step encoder.
The first stage of the encoder was a multilayer perceptron (MLP) with two layers of 128 units each and \texttt{LeakyReLU} activations, followed by a final projection to 8-dimensional ``information bottleneck'' space over $\tilde{U}$.
The second stage was another MLP with two layers of 128 \texttt{LeakyReLU} units, taking $\tilde{U}$ to the soft discretization space over $U$: a two dimensional space followed by a \texttt{softmax} activation.

In the information bottleneck space, transmitted information was penalized through the Kullback-Leibler (KL) divergence between the encoded conditional distributions $p(\tilde{u}|x)$ and the prior $r(\tilde{u})$, a standard normal distribution $\mathcal{N}(\boldsymbol{0}, \boldsymbol{1})$.
As is typical for variational autoencoders~\cite{vae}, the conditional distributions were parameterized as Gaussians with diagonal covariance matrices.
Values were sampled from the conditional distributions following the reparameterization trick~\cite{vae}, and the sampled values $\tilde{u}$ were passed through the second stage.
The \texttt{softmax} activation mapped embeddings to a probability distribution over two outcomes (representing assignment to a binary partition).
The temperature parameter in the \texttt{softmax} was taken to be one during training and changed to zero discontinuously after training (i.e., making the partition assignment an \texttt{arg max} operation on the probability vector).

The $L$ embeddings $u_1, ..., u_L$ were then concatenated and passed as input to the predictive model, $g(u_1, ..., u_L)$, an MLP consisting of two layers of 256 units with \texttt{LeakyReLU} activation, followed by a linear projection to 32 dimensions (i.e., a final dense layer with 32 units and no activation function).
The 32-dimensional space was shared with the output of a reference state encoder $h(x_\text{ref})$, another MLP with two layers of 256 units with \texttt{LeakyReLU} activation, followed by a linear projection to 32 dimensions.
In this 32-dimensional shared latent space the InfoNCE loss~\cite{oord2018InfoNCE} was evaluated by pairing up each length $L$ trajectory with its reference state, using the squared Euclidean distance as a measure of similarity between the embedding vectors.
The remainder of the batch size of 2048 provided negative samples for the InfoNCE loss.

To traverse the low information Pareto front, we used nonlinear IB~\cite{kolchinsky2019nonlinear}, squaring the information bottleneck penalty ($I(U_i;X_i)$) in the Lagrangian (Eqn.~\ref{eqn:dib} in the main text).
The loss was
\begin{equation}
    \mathcal{L}_\textnormal{VIB} = \mathcal{L}_\text{InfoNCE}(g(u_1,...,u_L),h(x_\text{ref})) + \beta \sum_i^L \left ( D_\textnormal{KL}(p(\tilde{u}_i|x_i)||r(\tilde{u})) \right)^2.
\end{equation}

We decreased $\beta$ in equally spaced logarithmic steps from $10$ to $10^{-4}$ over 20,000 steps, and stopped once the information per measurement $I(U;X)$ exceeded one bit (with $I(U;X)$ estimated as in Ref. \cite{dib2} using lower and upper bounds from \citet{poole2019variational}).
The Adam optimizer was used with a learning rate of $3\times10^{-4}$.

The low-dimensional continuous-valued states underwent \textit{positional encoding}, a term from natural language processing that has been used in certain computer vision contexts to facilitate learning on low-dimensional features. 
Before being passed to the MLP, inputs were mapped to $x \leftarrow (x, \sin \omega_1 x, \sin \omega_2 x, ... )$) with frequencies $\omega_k = 2^k$ where $k \in \{1, 2,...10\}$~\cite{dib_ml}.

After training concluded, there remained a difficulty around how to use the conditional distributions in the bottleneck space.  
The embeddings \textit{are} the distributions; we sampled from them during training thanks to the reparameterization trick, but sampling during the conversion to a partition would leave stochasticity in the assignment, i.e. $H(U|X){>}0$.
We can instead view the distribution in the information bottleneck space as an ensemble of hard partitions, and the transmitted information is the degree to which the downstream layers can ``pinpoint'' the member of the ensemble.
From this perspective it becomes sensible to sample a fixed number of noise vectors, representing as many members of the ensemble, and use those noise vectors to assign each $x$ to a $u$.
The majority assignment of each point is then used for the partition, sans stochasticity.

\subsection{Randomly sampled discrete measurements (Fig.~\ref{fig:partitions} of main text)}
To generate a random sample of discrete measurements, we initialized MLPs with 1, 2, and 3 layers of 64 units each, with \texttt{ReLU} or \texttt{tanh} activation. 
The weights and biases were sampled from a normal distribution with a mean of 0.05 and a standard deviation of 0.5.
The dimension of the input was 2, corresponding to the coordinates of the Ikeda map, and the output dimension was 2 or 4, corresponding to the alphabet size for the measurement (i.e., the number of colors).
The continuous-valued outputs, one for each input point, were discretized by selecting the dimension in the output with the largest absolute value.
20 networks were sampled for each configuration (number of layers, activation, alphabet size), for 240 in total.

\section{Appendix D: Iterates of partitions found in the paper}
We display in Fig.~\ref{fig:iterates} three forward and three reverse iterations of partitions for the Ikeda, \henon, and logistic maps studied in this work.
The optimized partitions for the \henon\ and logistic maps differ from previously reported generating partitions, with modifications that increase the entropy: 0.94 bits up from 0.91 bits for H{\'e}non, and 0.99 bits up from 0.82 bits for the logistic map.

\section{Appendix E: Citation Diversity Statement}
Science is a human endeavour and consequently vulnerable to many forms of bias; the responsible scientist identifies and mitigates such bias wherever possible.
Meta-analyses of research in multiple fields have measured significant bias in how research works are cited, to the detriment of scholars in minority groups~\citep{maliniak2013gender,caplar2017quantitative,chakravartty2018communicationsowhite,dion2018gendered,dworkin2020extent,teich2022citation}.
We use this space to amplify studies, perspectives, and tools that we found influential during the execution of this research~\citep{zurn2020citation,dworkin2020citing,zhou2020gender,budrikis2020growing}.
We sought to proactively consider choosing references that reflect the diversity of the field in thought, form of contribution, gender, race, ethnicity, and other factors. 
The gender balance of papers cited within this work was quantified using a combination of automated \href{https://gender-api.com}{\color{blue}gender-api.com} estimation and manual gender determination from authors’ publicly available pronouns.
By this measure (and excluding self-citations to the first and last authors of our current paper), the references of the main text contain 3\% woman(first)/woman(last), 7\% man/woman, 16\% woman/man, and 74\% man/man. 
This method is limited in that a) names, pronouns, and social media profiles used to construct the databases may not, in every case, be indicative of gender identity and b) it cannot account for intersex, non-binary, or transgender people.
We look forward to future work that could help us to better understand how to support equitable practices in science.

\end{document}